# Toward Transparent Sequence Models with Model-Based Tree Markov Model

Chan Hsu*, Wei-Chun Huang†, Jun-Ting Wu*, Chih-Yuan Li*, Yihuang Kang*
*Department of Information Management, National Sun Yat-sen University, Kaohsiung, Taiwan
†Department of Critical Care Medicine, Kaohsiung Veterans General Hospital, Kaohsiung, Taiwan
chanshsu@gmail.com, wchuanglulu@gmail.com, lydiazwu@gmail.com,
stellali1801@gmail.com, ykang@mis.nsysu.edu.tw

**Abstract**

*In this study, we address the interpretability issue in complex, black-box Machine Learning models applied to sequence data. We introduce the Model-Based tree Hidden Semi-Markov Model (MOB-HSMM), an inherently interpretable model aimed at detecting high mortality risk events and discovering hidden patterns associated with the mortality risk in Intensive Care Units (ICU). This model leverages knowledge distilled from Deep Neural Networks (DNN) to enhance predictive performance while offering clear explanations. Our experimental results indicate the improved performance of Model-Based trees (MOB trees) via employing LSTM for learning sequential patterns, which are then transferred to MOB trees. Integrating MOB trees with the Hidden Semi-Markov Model (HSMM) in the MOB-HSMM enables uncovering potential and explainable sequences using available information.*

**Keywords:** Interpretable machine learning, Machine learning interpretability, Hidden semi-markov models, Subgroup analysis, Temporal data mining, Process discovery

## 1. Introduction

Electronic Health Records (EHR) offer immense potential for deriving valuable insights into patients' healthcare. Researchers increasingly leverage Machine Learning (ML) algorithms to diagnose patients, predict patients' prognoses, identify symptom risk factors, and discover interactions between drugs and diseases using EHR data. The growing accessibility of high-performance computing resources enables the development of complex ML models capable of revealing intricate patterns from data and achieving high accuracy comparable to physicians [1]. However, these sophisticated models may produce correct outcomes for the wrong reasons–shortcut learning problems [2], and detecting such problems is challenging. Additionally, the tremendous number of internal parameters can render these models as so-called "black-box" models, making them hard to interpret. These concerns are critical, especially when these models support high-stakes decision-making tasks like clinical diagnoses. In these contexts, model interpretability is as important as prediction accuracy.

To address the issue of model interpretability, researchers have been developing explainer models to interpret such complex models [3], [4]. However, the fidelity of explainer models' explanations of black-box models' predictions remains questionable. These gaps could result in incorrect explanations, undermining trust in the explainer and original models. To tackle this issue and prevent potential misinterpretations of ML models, some researchers have proposed using inherently interpretable models instead of explaining black-box models [5]. These interpretable models can provide a more direct and transparent understanding of the factors driving their predictions without further explanations. Consequently, we aim to distill knowledge from unexplainable Deep Neural Networks (DNNs) and enhance interpretable ML algorithms to deliver accurate and explainable predictions.

We propose an interpretable ML algorithm called the Model-Based tree Hidden Semi-Markov Model (MOB-HSMM), a derivative of the Classification Tree Hidden Semi-Markov Model (CTHSMM) [6], for making inferences on sequence data. The proposed algorithm can adapt to various tasks by defining parametric models within tree nodes that help predicts future states or patterns based on existing information. The MOB-HSMM we propose is capable of the following:

1. Making inferences on sequence data while providing straightforward explanations of predictions,
2. Identifying hidden states (sequential patterns) to expose various relationships between predictors and outcomes,
3. Inferring possible state sequences given observed sequential activities of subjects.

The rest of this paper is structured as follows: The next section reviews related works and background on CTHSMM, MOB trees, and knowledge distillation, laying the groundwork for our proposed approach. Section 3 introduces the proposed model, which leverages deep neural networks' capabilities to construct more precise interpretable probabilistic models. Section 4 presents and discusses the experimental results of applying the proposed method to a real-world dataset. Finally, Section 5 concludes the paper, summarizing the proposed model's contributions.

## 2. Background

The rising accessibility of computing resources has led to a surge in the application of Deep Learning (DL) in Machine Learning (ML) algorithms [7]. Integrating Artificial Neural Networks (ANN) and representation learning enables DNN models to decipher complex and abstract patterns from data, surpassing traditional ML algorithms in handling sequence data [8]. However, the intricate architecture and various parameters of DNN models create difficulties in understanding these black-box models. Interpreting decision logic and predictions is vital for high-stakes tasks such as disease diagnosis. To overcome this interpretability challenge, explainer models, like Local Interpretable Model-Agnostic Explanations (LIME), have been designed to elucidate black-box models [9]. However, some researchers advocate for inherently interpretable models instead of relying on explanations for black-box models [5], [10]. They posit that explainer models might offer low fidelity in their explanations, as they cannot replicate the precise predictions of black-box models.

In response to the interpretability challenge, Liu et al. suggest improving the predictive performance of ML models by distilling knowledge from Long Short-Term Memory (LSTM) to eXtreme Gradient Boosting (XGBoost) and interpreting XGBoost with Shapley Additive exPlanations (SHAP) [11], [12]. Knowledge distillation entails student models emulating the predictions of teacher models, allowing them to gain additional insights beyond the original features [13]. This response-based knowledge distillation strategy can demystify black-box models and enhance interpretability.

Within this framework, we propose a novel approach known as the MOB-HSMM. Rather than using tree ensembles, we transfer the knowledge acquired from LSTM to a single MOB tree, which merges MOB trees and Hidden Semi-Markov Models (HSMM). MOB trees facilitate identifying interactions and relationships among independent variables and outcomes via recursive data partitioning, proving more effective in unearthing interactions than classification trees [14], [15]. Furthermore, MOB trees can be combined with various parametric models, such as Generalized Linear Mixed-effects Models (GLMM), to identify moderators from longitudinal data [16], [17]. By harnessing the strengths of MOB trees and HSMM, our approach aims to provide interpretable predictions and insights into trends based on existing data.

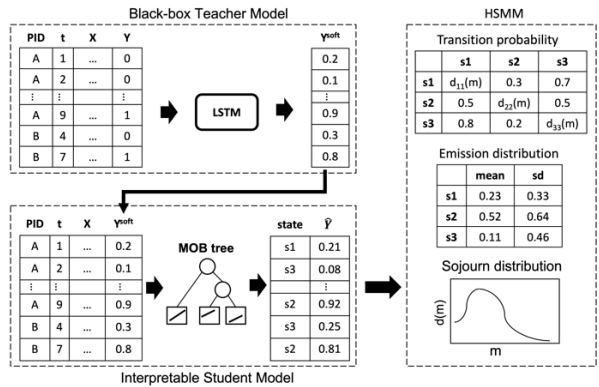

**Figure 1. The framework of the MOB-HSMM.**

The proposed approach is inspired by the Classification Tree Hidden Semi-Markov Models (CTHSMM) [6], which utilize tree models with learned rules to define the parameters of HSMM. In our model, the observed sequence represents the outcomes of predictive models, and the leaf nodes of the trees signify hidden states. This arrangement allows us to construct the transition matrix, emission distribution, and sojourn distribution by measuring the length of identical states, thereby discovering more defined and specific states from the data. The fusion of MOB trees and HSMM bolsters our ability to infer patients' future health states and augments the interpretability of the predictions.

## 3. Model-Based tree Hidden Semi-Markov Model

In this section, we illustrate our proposed approach, the Model-Based tree Hidden Semi-Markov Model (MOB-HSMM), along with a two-stage model learning process designed to uncover hidden states or patterns for corresponding observed sequences of outcomes: which are (1) the training of MOB trees using a teacher-student framework, where knowledge is distilled from a high-capacity teacher DNN, and (2) the construction of the HSMM using the MOB tree. The model learning process is shown in Figure 1.

Take an example of a typical data application that predicts the risk of death in the ICU. During the first stage of learning a black-box teacher model, we employ LSTM to convert the binary indicator of death into predicted probabilities indicative of the risk of death. We hypothesize that LSTM can discern informative patterns from sequential data and encode this knowledge into predicted probabilities. This knowledge is subsequently transferred to student MOB trees, empowering them to effectively model the relationship between independent variables and the risk of death. We assess the performance of both the teacher LSTM and student models using two metrics. The Area Under the Receiver Operating

Characteristic (AUROC) evaluates the extent of knowledge learned by LSTM. Cross-entropy between student models' predictions $\hat{y}$, and the LSTM's predicted probabilities $y^{soft}$ determines how effectively the student models can mimic LSTM. Upon selecting the optimal MOB tree, which exhibits the minimum cross-entropy loss, we proceed to the second stage and construct the HSMM with the selected MOB tree.

In the second stage, we initialize an HSMM using the MOB tree selected during the first stage. By partitioning data with similar patterns into distinct subspaces, MOB trees facilitate the correlation of each subspace with different relationships between predictors and outcomes. Consequently, we can define hidden states with the partition rules of the MOB tree and assign observations to the corresponding states. We then establish the transition matrix of the HSMM by counting state transitions in sequences within the training data. The emission distribution of each state is determined by the mean value and standard deviation of the state's outcomes, and each state's sojourn density functions are estimated using kernel density estimation with the Gaussian kernel.

Figure 2 shows an example of a MOB tree that partitions data with PEEP (positive end-expiratory pressure) and FiO2 (inspired fraction of oxygen) into three leaf nodes, each signifying a unique relationship between fluid balance and mortality risk. This MOB tree aids in identifying the differing relationships between fluid balance and the risk of mortality across various subgroups. The fluid balance exhibits a positive correlation with the risk of mortality in the right node but a negative correlation in the middle node. As shown in Figure 3, we can designate the corresponding states of each observation using the MOB tree, count the transitions from one state to another, and estimate the sojourn distribution with the duration. As we access the HSMM defined with MOB trees, we can further employ the HSMM inferring the most probable state sequence with the Viterbi algorithm or simply predict the following possible states based on the current state. In the next section, we illustrate the MOB-HSMM with a real-world dataset to provide a more comprehensive understanding of the proposed MOB-HSMM and its application in real-world scenarios.

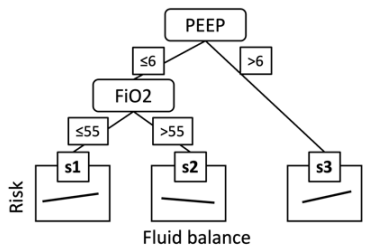

**Figure 2. A MOB tree with three states/nodes.**

| PEEP | FiO2 | Fluid balance (ml/h) | Risk | State | Duration |
|---|---|---|---|---|---|
| 7 | 60 | 0.16 | 0.52 | 3 | 4 |
| 5 | 50 | 0.03 | 0.03 | 1 | 3 |
| 4 | 35 | 0.02 | 0.02 | 1 | 6 |
| 5 | 60 | 0.08 | 0.05 | 2 | 3 |

**Figure 3. An example of ICU data with corresponding states.**

**Table 1. Variable summary**

| Variable Type | Variable[a] |
|---|---|
| Demographic characteristics | Age, gender, height, weight, BMI. |
| Vital signs | HR, ABPmean, BT_C. |
| Laboratory tests | PH, PO2, PCO3, HCO3, Na, K, Ca, lactate, glucose, BE, hematocrit, platelet, BNP, and CRP |
| Pivotal treatment | Indicators of ventilators (including invasive and non-invasive), PEEP, FiO2; indicators of Dopamine, Dobutamine, Epinephrine, Norepinephrine, and Vasopressin. |
| Fluid balance | 24h Urine output; fluid input, output, and balance. |

[a]Abbreviation: BMI body mass index, HR heart rate, ABPmean mean arterial blood pressure, BT_C body temperature in Celsius, BE base excess in the extracellular fluid compartment, BNP B-type natriuretic peptide, CRP C-reactive protein, PEEP positive end-expiratory pressure, FiO2 inspired fraction of oxygen

## 4. Experiment and Discussion

In this section, we demonstrate the application of the MOB-HSMM algorithm using a real-world dataset from the Kaohsiung Veterans General Hospital (KSVGH). The purpose is to predict mortality risk, explain the prediction, and identify hidden patterns or states corresponding to the risk. The dataset includes data from 1307 patients, 396 of whom died upon discharge from the ICU. It comprises 198,681 observations and numerous variables such as demographic details, vital signs, laboratory tests, pivotal treatments, and fluid balance. These variables are summarized in Table 1.

Health Information Systems (HIS) data often have missing values due to irregular recording and differing collection frequencies across variables. To maximize the utilization of available information, we used interpolation for accumulated variables and the Last Observation Carried Forward (LOCF) method to impute missing values for variables linked to vital signs and laboratory tests. We divided the patients randomly into two groups, alive or dead, and allocated 20% of patients from each group for testing while the remaining were used for training. This train-test split is depicted in Figure 4, consisting of 157,850 observations in the training set and 40,831 observations in the testing set.

The outcome variable "mortality" presents a significant class imbalance problem due to repeated patient recordings, with positive cases only appearing in a patient's final observations. To solve this problem, we applied repeated oversampling of positive observations in the training set, randomly cropping sequences of deceased patients at varying lengths, as demonstrated in Figure 5. This process elevated the positive observations ratio from 0.2% to 17.3%. Given the time-series nature of the data, we implemented an LSTM as the teacher model comprising three layers with 256, 64, and 8 neurons, respectively. We used the binary focal loss as the loss function to prioritize minor positive cases. The LSTM model achieved AUROCs of 0.98 and 0.75 for the training and testing sets, respectively. We postulate that the LSTM can identify important information about patients' recoverable critical events from the data. Consequently, we utilized the LSTM to infer the training and testing sets, transferring knowledge to student models.

We transformed the predicted probabilities into logits before training the student models to predict soft targets. This ensures that the students do not predict values outside the range of probabilities, which are constrained between 0 and 1. We employed GLMM trees to capture patient heterogeneity, combining GLMM and MOB trees. This allows us to uncover interactions among GLMMs and other partition variables of the trees. We defined the balance of intravenous fluid input and outflow as the fixed effect, given that intravenous fluid therapy is a common treatment for critically ill patients in the ICU [18].

To ascertain that the models have adequately learned from the data, we used the sliding window approach to assess the models' predictive capabilities [19]. This approach, a variant of the prequential block approach, can avoid the potential pitfall of training models with newer observations and subsequently evaluate them with older ones. The approach estimates model performance using the most recent data by depreciating earlier data folds. As a result, the prequential block sliding window approach can offer a more precise evaluation of model performance compared to traditional k-fold cross-validation. In our experiment, we divided each patient's observations into five folds, training the models using the first four folds and evaluating their performance using the next fold, as illustrated in Figure 6. We assessed the models' performance on the testing set for each fold and reported the average performance in Table 2.

Table 2 compares the predictive performance of a GLMM tree and other models, including Random Forest (RF), eXtreme Gradient Boosting (XGBoost), Classification And Regression Tree (CART), and Mixed-Effects Random Forest (MERF). The results suggest that GLMM trees can match the performance of ensemble models on training and validation sets. The comparison of cross-entropy among models indicates that GLMM trees can mimic the teacher LSTM as effectively as RF and XGBoost. This result implies that GLMM trees can adeptly use the mixed-effects design to learn patient differences.

However, the AUROC of GLMM trees and MERF on the testing set underscores a weakness of mixed-effect models: they struggle to predict outcomes accurately for new patients with no information. All experiments were conducted in R version 4.2.2 [20] with packages glmertree [17], rpart [21], ranger [22], mhsmm [23], and xgboost [24]; and Python 3.8 [25] with Keras [26] and TensorFlow 1.14 [27].

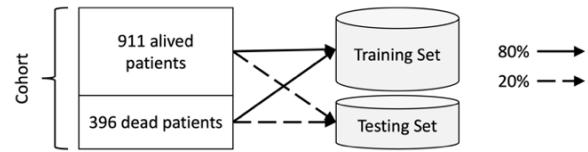

**Figure 4. Splitting observations into the training and testing sets according to the patient's mortality.**

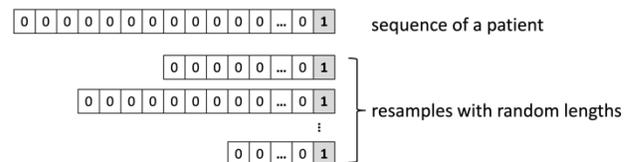

**Figure 5. An example of over-sampling on observations with the minor class.**

### Table 2. Performance comparison

| Model | Cross-Entropy | | | AUROC | | |
|---|---|---|---|---|---|---|
| | *Train* | *Valid* | *Test* | *Train* | *Valid* | *Test* |
| Random Forest (# of tree=500) | 0.1853 | 0.2174 | 0.2037 | 0.9152 | 0.8509 | 0.7079 |
| XGBoost (max_depth=4, eta=0.01, nround=6000) | 0.1789 | 0.2115 | 0.2081 | 0.9494 | 0.8878 | 0.7447 |
| CART (cp=0.001) | 0.1908 | 0.2231 | 0.2179 | 0.7626 | 0.7601 | 0.6200 |
| MERF (fluid balance as fixed effect, # of tree=500) | 0.1781 | 0.2134 | 0.2147 | 0.9438 | 0.8830 | 0.6999 |
| GLMM Tree (fluid balance as fixed effect, alpha=1e-30) | 0.1825 | 0.2079 | 0.2082 | 0.9259 | 0.8843 | 0.5603 |

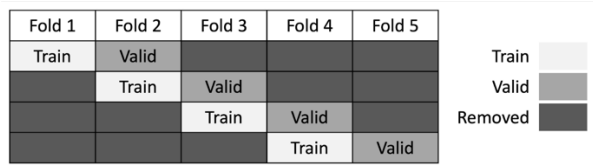

**Figure 6. Sliding window approach to estimate model performance.**

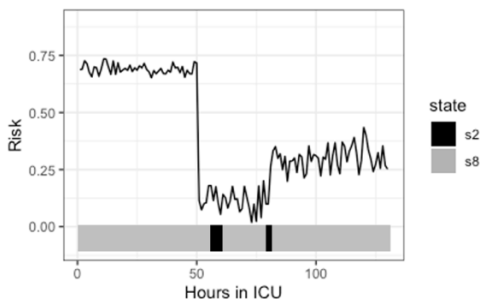

**Figure 7. An example of the most probable vital sign sequence given mortality risk within 150 hours.**

GLMM trees can be transformed into IF-THEN rules for interpretability. Table 3 outlines the rules of the GLMM tree discussed in Table 2. These rules elucidate the criteria of each subgroup and the relationships among partition variables, fluid balance, and mortality risk. For instance, the first rule suggests that a patient with low PEEP, no pivotal treatments, appropriate blood pressure, and balanced fluid input and outflow may have a lower mortality risk. Our method also allows us to predict the future health status of patients with HSMM or to identify possible shifts in health states using the Viterbi algorithm based on a sequence of mortality risks. Figure 7 graphically depicts a patient's potential state sequence, transitioning from high to lower risk and reverting to high risk. This fluctuating state sequence suggests that the patient, initially in state s8, briefly transitioned to state s2 before reverting to state 8. This sequence suggests that the patient initially received treatment with a higher PEEP and pivotal treatments like dopamine, briefly received treatment with lower PEEP and without pivotal treatment, and, as the risk escalated, reverted to treatment with higher PEEP and pivotal treatments. These potential health state inferences can assist clinicians in the early adjustment of their treatment plans, even though the projected state sequences may not perfectly match the actual states.

In our study, we posited the teacher model could detect patient mortality, recognize patterns of critical conditions, and identify high-risk periods during hospitalization. However, this assumption may not align with conventional metrics designed to measure models' predictive performance. As more high-risk periods are identified, the model's performance metrics decrease apart from death. Therefore, employing more suitable metrics to evaluate the model's ability to capture critical conditions other than the outcome could enhance the interpretability and reliability of the MOB-HSMM.

## 5. Conclusion

We introduce the Model-Based tree Hidden Semi-Markov Model (MOB-HSMM), an innovative and interpretable solution for predicting mortality risk in ICU patients while offering transparent explanations. Our approach leverages MOB trees and hidden semi-Markov models to predict patients' future health states. Using model-based recursive partitioning, we uncover latent states that reveal diverse relationships between independent variables and outcomes. Our results demonstrate that a single interpretable GLMM tree can perform comparably to other ensemble models by distilling knowledge from deep neural networks. Additionally, our approach is generalizable to different MOB tree algorithms. Most importantly, MOB-HSMM can offer insights into future scenarios based on available information by collaborating with Hidden Semi-Markov Models.

**Table 3. Observation matrix with state definition of the GLMM tree trained with fluid balance as fixed effects.**

| State | µY | Intercept | Coefficient | State Rule |
|---|---|---|---|---|
| s1 | 0.0147 | 0.0222 | -0.0608 | PEEP ≤ 6 & no pivotal treatments & Glucose ≤ 490 |
| s2 | 0.0143 | 0.0483 | 1.5669 | PEEP ≤ 6 & no pivotal treatments & Glucose > 490 & ABPmean ≤ 91 |
| s3 | 0.0086 | 0.1868 | -0.8007 | PEEP ≤ 6 & no pivotal treatments & Glucose > 490 & ABPmean > 91 |
| s4 | 0.0152 | 0.0110 | 0.4385 | PEEP ≤ 6 & have pivotal treatments & Lactate ≤ 8.8 |
| s5 | 0.0267 | 0.0148 | 3.0419 | PEEP ≤ 6 & have pivotal treatments & Lactate > 8.8 |
| s6 | 0.0269 | 0.0246 | -0.7774 | PEEP > 6 & no pivotal treatments & CRP ≤ 1.38 |
| s7 | 0.0306 | 0.0558 | 0.6934 | PEEP > 6 & no pivotal treatments & CRP > 1.38 |
| s8 | 0.0517 | 0.0103 | 2.2638 | PEEP > 6 & have pivotal treatments & SEX = 0 & weight ≤ 45.8 |
| s9 | 0.0363 | 0.0478 | 0.0494 | PEEP > 6 & have pivotal treatments & SEX = 0 & weight > 45.8 |
| s10 | 0.0339 | 0.0240 | 1.3682 | PEEP > 6 & have pivotal treatments & SEX = 1 |

## Acknowledgement

This work is part of the NSYSU and KSVGH Joint Research Project. The authors thank Dr. Chien-Wei Hsu and Dr. Mei-Tzu Wang for their valuable clinical advice.